\documentclass[twocolumn,a4paper]{arxiv}

\usepackage{dm-colors}
\usepackage{bbding}
\usepackage{enumitem}
\usepackage{graphicx}
\usepackage{csquotes}
\usepackage{multirow}
\usepackage[bibstyle=nature,citestyle=numeric-comp,%
            natbib=true,backend=biber,maxbibnames=99,%
            giveninits=false,sorting=none]{biblatex}
\usepackage[pagebackref=false,breaklinks=false,%
    colorlinks=true,bookmarks=true,citecolor=ourdarkblue,%
    urlcolor=ourdarkblue,linkcolor=ourdarkblue]{hyperref}
\usepackage{float}
\usepackage{dblfloatfix}
\usepackage{nicefrac}

\hypersetup{pdfauthor={Gioele Barabucci, Victor Shia, Eugene Chu, Benjamin Harack, and Nathan Fu},pdftitle={Combining Insights From Multiple Large Language Models Improves Diagnostic Accuracy}}

\newcommand{\MEAN}[2]{${#1\%\pm{}#2pp}$}
\newcommand{\VALUE}[1]{${#1}$}

\title{Combining Insights From\\
Multiple Large Language Models\\
Improves Diagnostic Accuracy}

\author[$\dagger$,1]{Gioele Barabucci}
\author[2,3]{Victor Shia}
\author[4]{Eugene Chu}
\author[3,5]{Benjamin Harack}
\author[3]{Nathan Fu}

\affil[1]{University of Cologne, }
\affil[2]{Harvey Mudd College, }
\affil[3]{The Human Diagnosis Project, \authorcr{}}
\affil[4]{Kaiser Permanente, }
\affil[5]{University of Oxford}

\renewcommand{\correspondingauthor}[1]{$\dagger$~Corresponding author: \url{gioele.barabucci@uni-koeln.de}}

\begin{document}

\begin{abstract}

\textbf{Background}
Large language models (LLMs) such as OpenAI's GPT-4 or Google's PaLM 2 are proposed as viable diagnostic support tools or even spoken of as replacements for ``curbside consults''.\newline
However, even LLMs specifically trained on medical topics may lack sufficient diagnostic accuracy for real-life applications.

\textbf{Methods}
Using collective intelligence methods and a dataset of 200 clinical vignettes of real-life cases, we assessed and compared the accuracy of differential diagnoses obtained by asking individual commercial LLMs (OpenAI GPT-4, Google PaLM 2, Cohere Command, Meta Llama 2) against the accuracy of differential diagnoses synthesized by aggregating responses from combinations of the same LLMs.

\textbf{Results}
We find that aggregating responses from multiple, various LLMs leads to more accurate differential diagnoses (average accuracy for 3 LLMs: \MEAN{75.3}{1.6}) compared to the differential diagnoses produced by single LLMs (average accuracy for single LLMs: \MEAN{59.0}{6.1}).

\textbf{Discussion}
The use of collective intelligence methods to synthesize differential diagnoses combining the responses of different LLMs achieves two of the necessary steps towards advancing acceptance of LLMs as a diagnostic support tool: (1) demonstrate high diagnostic accuracy and (2) eliminate dependence on a single commercial vendor.

\end{abstract}

\maketitle

\section{Background}

Large language models (LLMs) such as GPT-4 have been shown to be useful as support tools in various healthcare settings such as during tumor boards~\cite{GPT-Tumor-Board} or as a screening tool to match patient notes to best practice alerts \cite{LLM-Screening-Tool}. Their future use and deployment in healthcare is expected to parallel that of other AI tools such as automated electrocardiogram (ECG) anomaly detection, i.e., as support tools that provide insight to human practitioners to better inform their decisions \cite{NEJM-AI-Intro}.


In particular, there is ongoing research into the application of LLMs as summarization tools for patient and procedure information \cite{ChatGPT-discharge-summary} or as replacement for ``curbside consults'', especially in situations where colleagues may not be available (e.g., remote locations) or too expensive (e.g., underserved groups) \cite{NEJM-AI-Survey,LLM-ID-Consultation}.

Nevertheless, the use of LLM-based tools is impaired by their limited acceptance by medical professionals, among other factors \cite{AI-Ethics}.

One major factor driving this low rate of acceptance is lack of trust that an LLM can provide correct answers, and additionally, one that avoids so called ``hallucinations'', i.e., verisimilar but fictional responses. This lack of trust can in turn be traced back to lack of data on the performance (e.g., correctness, accuracy, specificity) of said LLM-based tools. Put bluntly, before trusting them, medical practitioners want to know: ``[Are they] good enough?'' \cite{NEJM-AI-Survey}.

Recent studies on the accuracy of LLMs have been shown them to be capable of performing well in certain medical contexts, but results seem to also vary depending on the research study or application, which paints an unclear situation. For instance, \citet{GPT-student-level} report that GPT-4 scores above 72\% of the readers of medical journals in 38 clinical case challenges. On the other hand, \citet{LLM-Pediatrics} report that GPT-4 has a diagnostic error rate of 83\% when confronted with 100 pediatric case challenges published on JAMA and NEJM.

This study investigates the use of collective intelligence methods to synthesize higher-accuracy differential diagnoses by aggregating differential diagnoses produced by a set of LLMs (even ones with low accuracy) in response to medical questions in the form of case vignettes.
Aggregating results from multiple LLMs could be the key to achieving high-accuracy responses (and potentially with fewer implausible or ``hallucinated'' responses). By employing algorithmic methods for knowledge aggregation from research on collective intelligence, it is possible to create a high-quality response to a question by aggregating lower-quality responses from multiple respondents \cite{CI-Methods}. For instance, past studies in the medical domain show that aggregating as few as three answers from inexperienced respondents led to high diagnostic accuracy (77\% accuracy), significantly above the performance of individual human experts (62.5\% accuracy) \cite{HDX-Barnett}.

\section{Methods}

This study can be summarized as follows: we sampled 200 clinical vignettes of real-life cases from the Human Diagnosis Project (Human~Dx) database, asked various LLMs to provide differential diagnoses for these cases, aggregated their responses using collective intelligence algorithms, and finally compared the accuracy of the individual LLM responses to the accuracy of the aggregated differentials.

The prompts and the Python scripts used to run the study are provided in the supplemental material. The case dataset and the LLM responses are available upon request.

\subsection{Dataset and case selection}

The data for this study is a set of 200 case vignettes, extracted from the Human~Dx database of clinical cases. Human~Dx is a multinational online platform in which physicians and medical students solve teaching cases, as well as offer clinical reasoning support to fellow users.

The 200 cases have been randomly sampled from the dataset used by \citet{HDX-Barnett}, restricting the sampling to text-only vignettes.

The correct diagnosis of each case (\emph{ground truth}) is known and has been validated by medical experts as part of the \citet{HDX-Barnett} study.

\subsection{Querying of LLMs}

Four general-purpose LLMs were asked to solve each case by providing a differential with five ranked diagnoses.
The four LLMs used in this study are: OpenAI GPT-4, Google PaLM 2 for text (text-bison), Cohere Command, Meta Llama 2 (llama-2-70b-f).

All prompts used to query the LLMs follow the same template: ``\emph{[CASE TEXT]} What is the differential (list format of common shorthand non-abbreviated diagnoses) for the above case? Respond with ONLY diagnosis names (one per line) up to a max of 5.'', where \emph{[CASE TEXT]} is replaced with the textual description of the case vignette.
The actual prompts vary slightly between different LLMs because of different query paradigms (e.g., GPT-4 uses a \emph{chat} paradigm while PaLM 2 uses a \emph{text generation} paradigm).

In order to obtain cleaner differential diagnoses for combining in collectives in the scoring process, a round of manual prompt engineering has been carried out using less than 5 Human~Dx case vignettes to help with the format and structure of the response.

We are highly confident that the case vignettes used in this study are not part of the training corpora of these LLMs. First, because the case vignettes are only available to users logged into the Human~Dx application and to select research partners. Second, because there are contractual agreements in place between Human~Dx and the providers of the LLMs that forbid the use of data included in prompts as training material (except for Cohere). Finally, the correct diagnoses were never included in any of the prompts to the various LLMs.

\begin{figure*}[hb!]
    \includegraphics[trim={0 24pt 0 0},width=\textwidth]{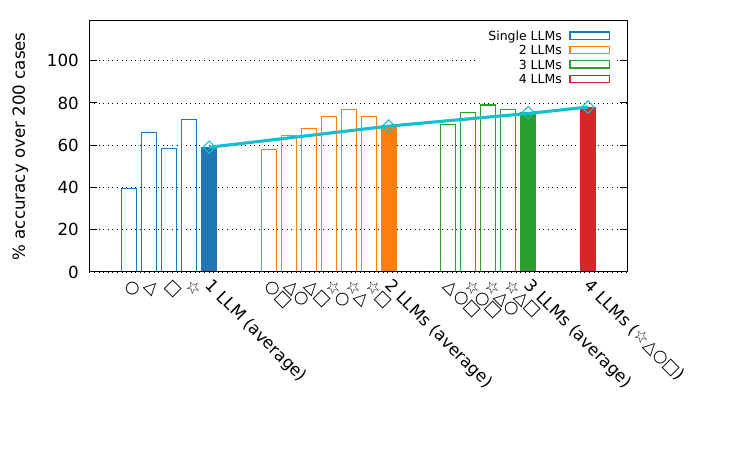}
    \caption{\label{fig:average-aggr-trend} Synthetic differential diagnoses aggregated from different LLMs show a greater diagnostic accuracy compared to differential diagnoses produced by single LLMs.
    This graph provides a visual representation of the data presented in Table~\ref{tab:top5-results}.\\
    $\bigcirc$ = Cohere Command, $\triangle$ = Google PaLM 2, $\square$ = Meta Llama 2, \raisebox{-0.5mm}{\FiveStarOpen} = OpenAI GPT-4}
\end{figure*}

\subsection{Collective intelligence and response aggregation}

Starting from the differential diagnoses provided by the single LLMs, 11 synthetic differential diagnoses have been generated by aggregating the single differential diagnoses in all possible combinations (6 two-fold combinations, 4 three-fold combinations, 1 four-fold combination).

The aggregation method is a frequency-based, \mbox{$1/r$-weighted} method similar to those used in other collective intelligence studies focused on diagnostic tasks via differential diagnoses \cite{HDX-Barnett, CROME}:

\begin{enumerate}

    \item \emph{Normalization}: All diagnoses in the differentials are normalized by removing common prefixes (e.g., ``syndrome'',  ``disorder''), stop words (e.g., ``by'', ``of'', ``with''), and punctuation signs. In addition, synonyms are merged into preferred terms, following the matching established by \citet{HDX-Barnett}.

    \item \emph{Extraction of unique diagnoses}: The set of all unique normalized diagnoses present across all differentials is created.

    \item \emph{$1/r$ weighting}: Inside each differential each diagnosis is given an \emph{individual score} calculated as the inverse of the rank \emph{r} of the diagnosis in the differential (i.e., the first diagnosis is given the score $\nicefrac{1}{1} = 1$, the second $\nicefrac{1}{2} = 0.5$, the third $\nicefrac{1}{3} = 0.33$, etc).

    \item \emph{Aggregation}: Each of the unique diagnoses in the set created in step 2 is given a \emph{aggregate score} calculated by adding all the individual scores of that diagnosis across all differentials.

    \item \emph{Synthesis}: A synthetic differential is generated by taking the five unique diagnoses with the highest aggregate score and ranking them by their score in decreasing order.

\end{enumerate}

\subsection{Accuracy measure}

The accuracy of a solver (either an LLM or a group of LLMs) is calculated as the percentage of correctly diagnosed cases among all cases.

For this study we consider a case to be correctly diagnosed by a solver if the differential provided by that solver for that case contains the correct diagnosis among the five highest ranked diagnoses.

This so-called \emph{TOP-5 matching} mirrors similar correctness measures used in previous studies \cite{HDX-Barnett, CROME}. Results obtained using TOP-1 or TOP-3 matching are provided in the supplemental material.

\begin{figure*}[ht!]
\includegraphics[trim={0 36pt 0 36pt},width=\textwidth]{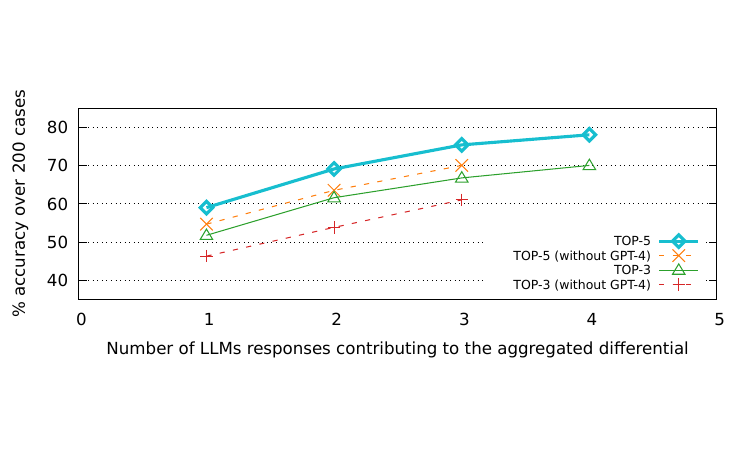}
    \caption{\label{fig:top-x-trends} Increasing the number of LLMs contributing to a synthetic differential leads to an increase in accuracy also (a) when the definition of correctly diagnosed case is made stricter by considering only the 3 highest ranked diagnosis in a differential (TOP-3 matching) and (b) when the top-performing LLM, GPT-4, is excluded from the experiment.}
\end{figure*}

\begin{table*}[hb!]
\begin{singlespace}
{\footnotesize
\newcommand{\s}{\hphantom{S}}
\begin{tabular*}{0.995\textwidth}{@{}c@{\extracolsep{\fill}}lcc}
\toprule
\s Group size & LLMs in group      & Accuracy & Average accuracy \\
\midrule
\s 1 & Cohere Command              & \VALUE{39.5\%} & \multirow{4}{*}{\MEAN{59.0}{6.1}} \\
\s 1 & Google PaLM 2                & \VALUE{66.0\%} & \\
\s 1 & Meta Llama 2                 & \VALUE{58.5\%} & \\
\s 1 & OpenAI GPT-4                & \VALUE{72.0\%} & \\
\midrule
\s 2 & Cohere Command, Meta Llama 2  & \VALUE{58.0\%} & \multirow{6}{*}{\MEAN{69.1}{2.6}} \\
\s 2 & Google PaLM 2, Cohere Command & \VALUE{64.5\%} & \\
\s 2 & Google PaLM 2, Meta Llama 2    & \VALUE{68.0\%} & \\
\s 2 & OpenAI GPT-4, Cohere Command & \VALUE{73.5\%} & \\
\s 2 & OpenAI GPT-4, Google PaLM 2   & \VALUE{77.0\%} & \\
\s 2 & OpenAI GPT-4, Meta Llama 2    & \VALUE{73.5\%} & \\
\midrule
\s 3 & Google PaLM 2, Cohere Command, Meta Llama 2 & \VALUE{70.0\%} & \multirow{4}{*}{\MEAN{75.3}{1.6}} \\
\s 3 & OpenAI GPT-4, Cohere Command, Meta Llama 2 & \VALUE{75.5\%} & \\
\s 3 & OpenAI GPT-4, Google PaLM 2, Cohere Command & \VALUE{79.0\%} & \\
\s 3 & OpenAI GPT-4, Google PaLM 2, Meta Llama 2 & \VALUE{77.0\%} & \\
\midrule
\s 4 & Google PaLM 2, Cohere Command, Meta Llama 2, OpenAI GPT-4 & \VALUE{78.0\%} & \MEAN{78.0}{0.1} \\
\bottomrule
\end{tabular*}
}
\end{singlespace}
\caption{\label{tab:top5-results}Diagnostic accuracy of single LLM and groups of LLMs over the 200 cases present in the dataset. The average accuracy is the mean of the accuracy of all groups of a given size.} 
\end{table*}

\section{Results}

The main finding of this study is that the accuracy of differential diagnoses created by aggregating differential diagnoses from multiple LLMs using collective intelligence methods (accuracy for 3 LLMs: \MEAN{75.3}{1.6}) are consistently better than the accuracy of differential diagnoses produced by single LLMs (average accuracy of single LLMs: \MEAN{59.0}{6.1}).

The average accuracy of individual LLMs is \MEAN{59.0}{6.1}, i.e., on average a LLM produces a ranked differential diagnosis that contains the right diagnosis in \VALUE{59.0\%} of the cases.
The average accuracy of groups of LLMs increases as the group size grows: the average accuracy for groups of 2 LLMs is \MEAN{69.1}{2.6}, for 3 LLMs is \MEAN{75.3}{1.6}, for 4 LLMs is \MEAN{78.0}{0.1}.
Figure~\ref{fig:average-aggr-trend} shows the individual and average accuracy of the single LLMs and of groups of LLMs.

This finding holds true also when the definition of correctly diagnosed case is made stricter by considering only the 3 highest ranked diagnosis in a differential (TOP-3 matching), as illustrated in Figure~\ref{fig:top-x-trends}.

This trend is also confirmed when the differential diagnosis of the LLM with the highest individual accuracy, GPT-4, is excluded from the experiment (average accuracy of single LLMs: \MEAN{54.6}{6.4}, of groups of 2 LLMs: \MEAN{63.5}{2.3}, 3 LLMs: \MEAN{70.0}{0.1}).

The supplemental material provides data on these alternative evaluation methods.

\section{Discussion}

This study demonstrates the feasibility and validity of using collective intelligence methods to combine low-accuracy differentials from multiple LLMs into synthetic high-accuracy differentials. The degree of increase in accuracy achieved by the method employed in this study is in line with similar results in the field of collective intelligence, both in the medical field \cite{HDX-Barnett, CROME} and outside \cite{CI-group-lie-detection,CI-Methods}.

The mechanism that allows this increase in accuracy is that the presented aggregation method emphasizes plausible diagnoses (likely to be present in the differential returned by multiple LLMs, and so, bound to have a higher aggregate score), while minimizing the effects of hallucinated diagnoses (likely to be present in only one of the LLMs). This can be seen as an instance of the \emph{Anna Karenina principle} (good answers are common to many LLMs, bad answers are local to a specific LLM).

This principle is exploited by similar techniques used in LLM research such as ensemble methods \cite{LLM-Blender, ONE-LLM} or multi-agent debates \cite{ChatEval-Debate}.

Two factors differentiate the method employed in this study from these techniques: universality and simplicity.
First, this method works despite using LLMs with varying querying approaches and technical differences, and can thus be easily extended to work with any combination of LLMs.
Second, the simplicity of this method means that not only can it be easily integrated in existing software applications, but could even be performed by medical personnel manually querying separate LLMs and synthesizing the results themselves.

The trust of medical practitioners in LLM-based tools could be strengthened by the application of aggregation methods like the one employed in this study. In particular, knowing that multiple sources contributed would increase clinician confidence in the final differential and lessen the fear of having encountered one of the many mistaken answers or hallucinations that LLMs are known to produce~\cite{NEJM-AI-Survey}.

An additional advantage of the use of knowledge aggregation methods is preventing vendor lock-in, removing the need to engage with a single, potentially expensive, or legally problematic LLM vendor in order to obtain high-quality diagnostic differentials.
The use of aggregation methods like the one we propose would address these issues by enabling the use of multiple cheaper LLMs, or alternatively, locally-deployed and fine-tuned LLMs.
For instance, our results show that the 3-LLM group without GPT-4 (the top-performing LLM) offers a diagnostic accuracy within a couple of percentage points of GPT-4 alone.

\balance

With a clear baseline on diagnostic accuracy and trust, LLM-based tools can become valuable support instruments that can speed up diagnosis, reduce diagnostic mistakes and costs, and provide additional consulting services in underserved areas.

While this study shows that aggregating differentials produced by current LLMs leads to improved diagnostic accuracy, further studies are needed to examine the impact of future LLMs specialized on medical topics or the use of participatory AI methods (for instance, the synthesis of differentials aggregating responses from both LLMs and human practitioners).

\section{Acknowledgements}

The authors would like to thank Nikolas Zöller of the Max Plank Institute for Human Development for his valuable and constructive feedback.

The authors would also like to thank Irving Lin and Jay Komarneni of the Human Diagnosis Project for their suggested framing and review.

This work is supported by the European Union’s Horizon Europe Research and Innovation Programme under grant agreement No 101070588 (HACID: 
Hybrid Human Artificial Collective Intelligence in Open-Ended Domains).

\printbibliography

\end{document}